\documentclass[
]{ceurart}

\usepackage{listings}
\newcommand{\smallpm}[1]{\ensuremath{\,\pm\,#1}}
\newcommand{\bestval}[2]{\textbf{#1}\smallpm{#2}}
\begin{document}

%%
%% Rights management information.
%% CC-BY is default license.
\copyrightyear{2026}
\copyrightclause{Copyright for this paper by its authors.
  Use permitted under Creative Commons License Attribution 4.0
  International (CC BY 4.0).}

%%
%% This command is for the conference information
\conference{LaCATODA 2026: The 10th Linguistic and Cognitive Approaches to Dialog Agents Workshop at the 40th AAAI conference, January 20–27, 2026, Singapore Expo}

%%
%% The "title" command
\title{Partial Reasoning in Language Models: Search and Refinement Guided by Uncertainty}

%%
%% The "author" command and its associated commands are used to define
%% the authors and their affiliations.
\author[1,2]{Murilo da Luz}[%
email=muriloluz@ufg.br
]
\cormark[1]

\author[1,2]{Bruno Brandão}
\author[1,2]{Luana Martins}
\author[1,2]{Gustavo Oliveira}
\author[1,2]{Bryan de Oliveira}
\author[1,3]{Luckeciano Melo}
\author[1,2]{Telma Soares}

\address[1]{Advanced Knowledge Center for Immersive Technologies (AKCIT)}
\address[2]{Federal University of Goiás, Brazil}
\address[3]{OATML, University of Oxford}

%% Footnotes
\cortext[1]{Corresponding author.}

%%
%% The abstract is a short summary of the work to be presented in the
%% article.
\begin{abstract}
The use of Large Language Models (LLMs) for reasoning and planning tasks has drawn
increasing attention in Artificial Intelligence research. Despite their remarkable progress,
these models still exhibit limitations in multi-step inference scenarios, particularly in
mathematical and logical reasoning. We introduce PREGU (Partial Reasoning Guided by Uncertainty). PREGU monitors the entropy of the output distribution
during autoregressive generation and halts the process whenever entropy exceeds a defined threshold, signaling uncertainty. From that point, a localized search is performed in
the latent space to refine the partial reasoning and select the most coherent answer, using
the Soft Reasoning method. Experiments conducted with LLaMA-3-8B, Mistral-7B, and
Qwen2-7B across four reasoning benchmarks (GSM8K, GSM-Hard, SVAMP, and StrategyQA) showed performance greater than or similar to Soft Reasoning, indicating that entropy can serve as an effective signal to trigger selective refinement during reasoning.
\end{abstract}

%%
%% Keywords. The author(s) should pick words that accurately describe
%% the work being presented. Separate the keywords with commas.

%%
%% This command processes the author and affiliation and title
%% information and builds the first part of the formatted document.
\maketitle

\section{Introduction}

Large Language Models (LLMs) have demonstrated notable reasoning capabilities, often enhanced by techniques like Chain-of-Thought (CoT) prompting \cite{wang2023understandingchainofthoughtpromptingempirical}. However, their performance often falters in tasks demanding complex, multi-step structured reasoning, such as advanced mathematics or contextual planning. This limitation stems from the challenge LLMs face in simulating long-term outcomes and exploring alternative reasoning paths, a process humans manage effectively \cite{hao2023reasoninglanguagemodelplanning}.

Traditional reasoning strategies often operate in the vast space of tokens, which can be inefficient due to the immense number of possible token combinations. An alternative is performing search in the latent space, where representations are more abstract, compressed, and lower in dimensionality, avoiding successive natural language encoding and decoding.

The Soft Reasoning (SR) \cite{zhu2025softreasoningnavigatingsolution} method recently introduced exploration in the latent space by iteratively adjusting the input embedding (specifically, the first token) to influence the subsequent deterministic text generation. While effective, SR's efficacy is limited by its strong reliance on the starting point—optimizing the latent space only from the initial prompt. Uncertainty in LLMs is dynamic, typically manifesting in intermediate steps when the model must integrate facts or choose between plausible hypotheses. Restricting optimization to the beginning of the sequence risks focusing computational effort on regions where the model is already confident.

This research proposes PREGU (Figure~\ref{fig:fig-solucao.png}), an adaptive extension of Soft Reasoning. PREGU dynamically identifies points of uncertainty during generation and triggers a focused search in the latent space only at those specific critical junctures.

\section{Background}
\subsection{Latent Space and Soft Reasoning}

The latent space (or embedding space in the context of Large Language Models) is an abstract, multi-dimensional mathematical representation of input data (such as words or phrases) that encodes semantic, syntactic, and contextual information. Operating within this space allows LLM agents to function more efficiently by using compressed and abstract representations of environmental dynamics.

Soft Reasoning performs a search by applying controlled Gaussian perturbations ($\sigma \epsilon_i$) to the embedding of the first token, $z$. This exploration is guided by Bayesian Optimization using the \textit{Expected Improvement} (EI) acquisition function \cite{frazier2018tutorialbayesianoptimization}. The quality of the generated sequence ($y$), derived from a perturbed embedding ($x$), is evaluated through a reward function defined as:
\begin{equation}
\label{eq:reward}
    f(x) = r_{\text{verifier}}(y) + r_{\text{coherence}}(y) \text{,}
\end{equation}

\noindent
where $r_{\text{verifier}}$ determines correctness — typically by employing the LLM itself as a black-box verifier through the \textit{Multi-Generate} approach \cite{zhu2025softreasoningnavigatingsolution} — and $r_{\text{coherence}}$ assesses the semantic and syntactic fluency of the generated text. SR thus enables the model to refine its reasoning efficiently by exploring smooth variations in generation trajectories within the latent space.

\subsection{Entropy as a Measure of Uncertainty}

\begin{figure*}[!t]
    \centering
    \includegraphics[width=1\linewidth]{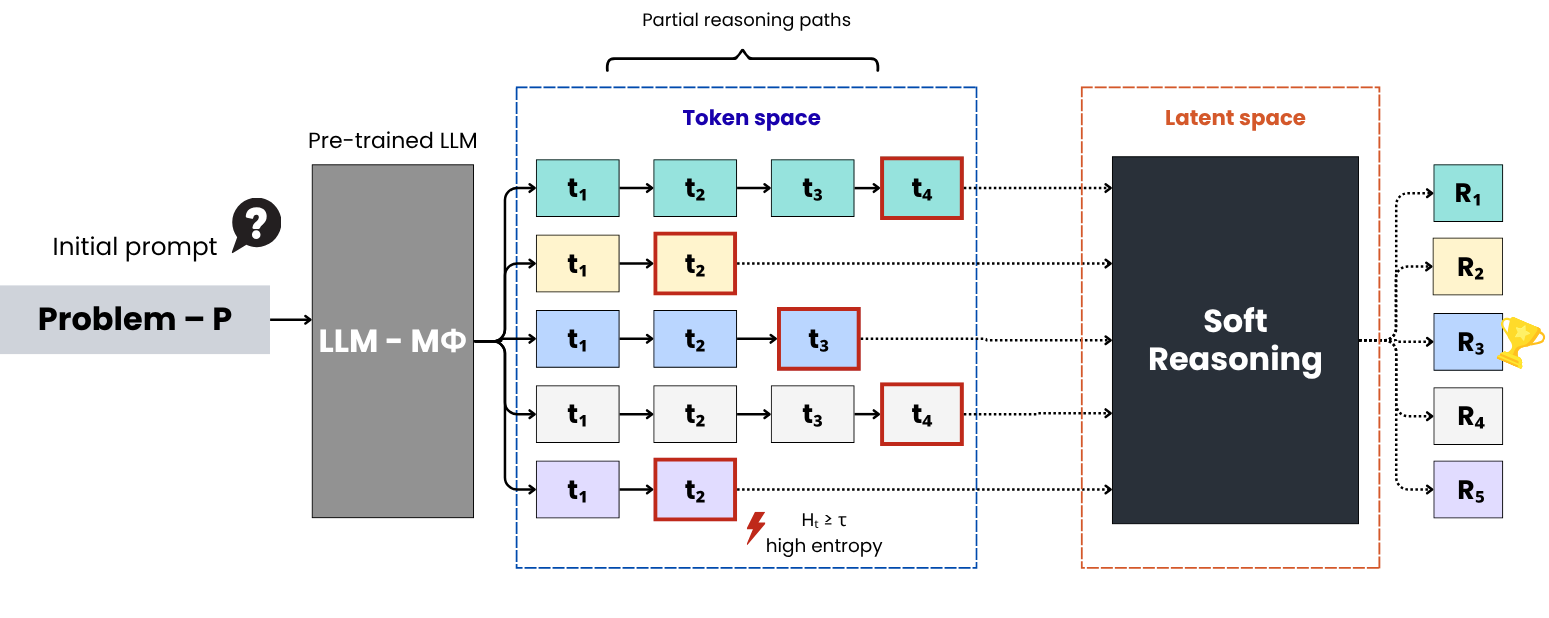}
    \caption{Overview of the PREGU architecture. The pre-trained LLM ($M_{\phi}$) generates multiple partial reasoning paths in the token space. When the entropy of the output distribution ($H_t$) exceeds the threshold ($\tau$), the generation halts, marking a region of uncertainty. Each partial reasoning sequence is then refined in the latent space using the Soft Reasoning method, producing candidate responses ($R_1, R_2, \ldots, R_n$), from which the highest-reward answer is selected.
}
    \label{fig:fig-solucao.png} 
\end{figure*}

A standard method for quantifying an LLM's internal uncertainty is by analyzing the probability distribution of its output tokens. The metric employed here is the \textit{Shannon Entropy} ($H_t$), defined as:
\begin{equation}
    H_t = - \sum_{j=1}^{K} p^{(j)}_t \log_2 p^{(j)}_t \text{.}
\end{equation}

A higher entropy value signifies a flatter or more uniform probability distribution, indicating greater ambiguity or indecision in selecting the next token. By setting an entropy threshold ($\tau$), the model's autoregressive process can be interrupted when high uncertainty is detected, thereby activating an auxiliary reasoning mechanism. This approach mimics human dual cognition models (\textit{System 1}: fast/intuitive; \textit{System 2}: slow/analytical), where a slower, analytical process is triggered only when intuition encounters an impasse or high doubt \cite{kahneman_thinking_2012}.

Inspired by dual-process theories of cognition, PREGU mirrors the interaction between intuitive and analytical reasoning systems: a fast, heuristic mode during confident segments and a slower, deliberative intervention when uncertainty peaks. In this view, the entropy threshold $\tau$ acts as a computational analogue of metacognitive control, triggering a shift from automatic generation to reflective reasoning.

\section{The PREGU Method}

Unlike prompt-based diversification strategies such as Chain-of-Thought \cite{wei2023chainofthoughtpromptingelicitsreasoning}, Self-Consistency \cite{wang2023selfconsistencyimproveschainthought}, ReAct \cite{yao2023reactsynergizingreasoningacting}, or Tree of Thoughts \cite{yao2023treethoughtsdeliberateproblem}, which primarily vary surface-level text to elicit reasoning, PREGU reallocates computation toward points of uncertainty detected via Shannon entropy. This direction follows latent-space reasoning explored by Soft Reasoning \cite{zhu2025softreasoningnavigatingsolution}, emphasizing optimization over internal representations rather than prompt reformulation.

\textit{PREGU} (\textbf{Partial Reasoning Guided by Uncertainty}) is designed to overcome the limitation of Soft Reasoning (SR), which optimizes only the initial token embedding. PREGU integrates entropy-based detection with latent space refinement in a two-stage process (Figure~\ref{fig:fig-solucao.png}).

\subsubsection*{Stage 1: Partial Reasoning Generation (Token Space)}
We denote by $M_{\phi}$ a pretrained language model with parameters $\phi$.
The language model $M_{\phi}$ performs autoregressive generation to produce $N$ candidate reasoning paths.
At each decoding step $t$, we monitor the Shannon entropy $H_t$ of the next-token distribution.
For each path, generation is interrupted \emph{at most once}, exactly at the \emph{first} token whose entropy crosses the threshold $\tau$ (i.e., the first $t$ such that $H_t \ge \tau$), yielding a partial reasoning prefix.

Entropy is estimated over the top-$K$ tokens (with $K=50$), balancing computational cost and representativeness; very low-probability tokens contribute minimally to the total uncertainty.

\paragraph{Minimum Prefix ($t_{\min}$):}the model must generate at least $t_{\min}$ tokens before uncertainty detection is allowed, ensuring sufficient semantic context.

\subsubsection*{Stage 2: Refinement (Latent Space)}
Each partial reasoning sequence generated in Stage~1 is treated as an extended prompt root ($p_i$).  
This partial reasoning is then refined independently using the Soft Reasoning method. The search is focused on the latent space starting from the point of interruption, allowing Bayesian Optimization to specifically address the ambiguity that triggered the halt.

The process yields multiple candidate answers $(R_1, R_2, \ldots, R_n)$, and the final solution ($a^{*}$) is selected based on the highest reward computed by the equation \ref{eq:reward}.

This structure combines \textit{breadth exploration} (through multiple partial reasoning paths) with \textit{focused depth exploration} (via latent space refinement), enabling a more adaptive and uncertainty-aware reasoning process.

\section{Experimental Setup}

\subsection{Configuration}

Experiments were conducted using three intermediate-scale (7–8 billion parameter) open-source LLMs under a zero-shot setting: \textbf{LLaMA-3-8B} \cite{grattafiori2024llama3herdmodels}, \textbf{Mistral-7B} \cite{jiang2023mistral7b}, and \textbf{Qwen2-7B} \cite{yang2024qwen2technicalreport}. These models present distinct architectural characteristics: LLaMA-3 emphasizes generalization, Mistral focuses on architectural efficiency through \textit{Sliding Window Attention} (SWA), and Qwen2 prioritizes robustness in mathematical reasoning tasks.

\paragraph{Benchmarks.}
The methodology was evaluated against four reasoning benchmarks:
\begin{itemize}
    \item \textbf{GSM8K:} Grade-school mathematical word problems  \cite{cobbe2021trainingverifierssolvemath}.
    \item \textbf{GSM-Hard:} A more challenging variant of GSM8K, requiring multi-step reasoning and compositional inference \cite{gao2022pal}.
    \item \textbf{SVAMP:} Tests semantic robustness by introducing superficial structural variations in mathematical problems \cite{patel2021nlpmodelsreallyable}.
    \item \textbf{StrategyQA:} Evaluates strategic and commonsense reasoning, requiring the decomposition of implicit sub-hypotheses \cite{geva2021didaristotleuselaptop}.
\end{itemize}

\paragraph{Baselines.}

We compare PREGU and Soft Reasoning against standard prompting and decoding baselines on GSM8K, GSM-Hard, SVAMP, and StrategyQA. \textbf{CoT} (Chain-of-Thought) encourages step-by-step reasoning by prompting the model to generate intermediate rationale before the final answer \cite{kojima2023largelanguagemodelszeroshot}. \textbf{SC} (Self-Consistency) samples multiple CoT solutions (e.g., using different temperatures) and selects the final answer by consensus (majority vote) \cite{wang2023selfconsistencyimproveschainthought}. \textbf{FIRE} increases diversity by applying a high temperature only to the first generated token while decoding subsequent tokens with regular execution sampling \cite{chen2025flaminghotinitiationregularexecution}. \textbf{CoT-Decoding} further diversifies reasoning by starting generation from the top-$k$ most likely first tokens and decoding a completion from each start \cite{wang2024chainofthoughtreasoningprompting}. For these baselines, we adopt the same evaluation protocol and report the baseline results as presented in the Soft Reasoning study \cite{zhu2025softreasoningnavigatingsolution}.

\begin{table*}[b]
\centering
\caption{Average accuracy (\%) — mean $\pm$ std over five runs. Bold marks best per benchmark.}
\label{tab:pregu-results}
\setlength{\tabcolsep}{4pt}   % compacta colunas sem reduzir fonte
\renewcommand{\arraystretch}{1.12} % levemente mais alto para legibilidade
\begin{tabular}{l l c c c c}
\toprule
\multirow{2}{*}{\textbf{Model}} & \multirow{2}{*}{\textbf{Method}}
& \textbf{GSM8K} & \textbf{GSM-Hard} & \textbf{SVAMP} & \textbf{StrategyQA} \\
\cmidrule(lr){3-6}
& & \textit{Zero-Shot} & \textit{Zero-Shot} & \textit{Zero-Shot} & \textit{Zero-Shot} \\
\midrule

\multirow{9}{*}{\textbf{LLaMA-3-8B}}
& CoT             & $53.0\smallpm{0.0}$ & $14.0\smallpm{0.0}$ & $61.0\smallpm{0.0}$ & $58.5\smallpm{0.0}$ \\
& SC($T{=}0.4$) & $73.0\smallpm{1.6}$ & $25.7\smallpm{0.4}$ & $79.1\smallpm{1.2}$ & $64.7\smallpm{0.7}$ \\
& SC($T{=}0.6$) & $73.6\smallpm{2.5}$ & $24.5\smallpm{1.1}$ & $76.1\smallpm{3.9}$ & $59.9\smallpm{2.0}$ \\
& SC($T{=}0.8$) & $65.0\smallpm{2.0}$ & $21.8\smallpm{1.3}$ & $69.6\smallpm{2.0}$ & $54.4\smallpm{2.6}$ \\
& FIRE             & $73.8\smallpm{2.3}$ & $25.2\smallpm{3.0}$ & $81.5\smallpm{0.8}$ & $63.0\smallpm{3.7}$ \\
& CoT-Decoding     & $73.9\smallpm{1.9}$ & $24.8\smallpm{1.3}$ & $83.2\smallpm{1.2}$ & $64.6\smallpm{1.6}$ \\
& Soft Reasoning   & $79.4\smallpm{1.2}$ & $28.2\smallpm{1.8}$ & \bestval{88.2}{1.3} & $67.2\smallpm{0.7}$ \\
& \textbf{PREGU (ours)} & \bestval{82.6}{1.6} & \bestval{35.2}{3.4} & $87.4\smallpm{0.9}$ & \bestval{68.6}{0.7} \\
\midrule

\multirow{9}{*}{\textbf{Qwen2-7B}}
& CoT             & $64.5\smallpm{0.0}$ & $40.0\smallpm{0.0}$ & $43.5\smallpm{0.0}$ & $63.0\smallpm{0.0}$ \\
& SC($T{=}0.4$) & $81.2\smallpm{0.6}$ & $47.5\smallpm{1.4}$ & $72.3\smallpm{2.0}$ & $67.1\smallpm{1.5}$ \\
& SC($T{=}0.6$) & $80.2\smallpm{1.9}$ & $46.2\smallpm{1.9}$ & $77.3\smallpm{1.2}$ & $67.5\smallpm{0.7}$ \\
& SC($T{=}0.8$) & $80.0\smallpm{0.9}$ & $47.3\smallpm{1.3}$ & $78.6\smallpm{2.1}$ & $67.0\smallpm{1.0}$ \\

\multirow{9}{*}{\textbf{Mistral-7B}}
& CoT             & $42.0\smallpm{0.0}$ & $14.5\smallpm{0.0}$ & $52.0\smallpm{0.0}$ & $62.0\smallpm{0.0}$ \\
& SC($T{=}0.4$) & $52.9\smallpm{0.5}$ & $19.5\smallpm{1.0}$ & $67.4\smallpm{2.5}$ & $69.3\smallpm{1.5}$ \\
& SC($T{=}0.6$) & $55.1\smallpm{3.6}$ & $20.7\smallpm{1.5}$ & $69.7\smallpm{1.6}$ & $64.2\smallpm{1.0}$ \\
& SC($T{=}0.8$) & $50.2\smallpm{2.6}$ & $19.1\smallpm{2.0}$ & $68.3\smallpm{0.9}$ & $64.9\smallpm{1.0}$ \\
& FIRE             & $47.2\smallpm{2.9}$ & $18.1\smallpm{1.9}$ & $67.1\smallpm{1.9}$ & $64.2\smallpm{1.0}$ \\
& CoT-Decoding     & $47.3\smallpm{3.0}$ & $16.6\smallpm{0.7}$ & $69.4\smallpm{2.5}$ & $63.5\smallpm{1.5}$ \\
& Soft Reasoning   & $61.4\smallpm{2.5}$ & $25.8\smallpm{1.8}$ & $72.2\smallpm{2.2}$ & $66.1\smallpm{1.9}$ \\
& \textbf{PREGU (ours)} & \bestval{66.0}{1.9} & \bestval{32.7}{1.4} & \bestval{74.8}{0.8} & \bestval{68.1}{1.9} \\
\bottomrule
\end{tabular}
\end{table*}

\paragraph{Hyperparameters.}
PREGU was evaluated using a standard hyperparameter configuration informed by preliminary studies of token-level entropy (see Sec.~\ref{sec:uncertainty}), including:
\begin{itemize}
    \item Entropy threshold: $\tau = 3.0$ bits
    \item Token sample size for entropy estimation: $K = 50$
    \item Search width: $N = 5$ partial reasoning paths
    \item \textit{Soft Reasoning} parameters: $k = 5$ samples, projected latent dimension $d = 50$
\end{itemize}

\section{Results}

The results in Table~\ref{tab:pregu-results} indicate that PREGU generally matches or improves upon Soft Reasoning across the evaluated benchmarks. We observe small regressions in isolated settings (e.g., Qwen-2-7B on GSM8K and LLaMA-3-8B on SVAMP), while maintaining gains on GSM-Hard, StrategyQA, and most remaining cases. Across five runs, the method remained stable and did not exhibit systematic performance degradation, suggesting that combining entropy-guided interruption with localized latent-space refinement may improve reasoning consistency across domains, especially on multi-step inference chains such as GSM-Hard. 

Notably, performance on the GSM-Hard dataset improved, where PREGU showed robustness on long, multi-step inference chains. Even with the search being initiated only at the first point of uncertainty, the method successfully refined reasoning trajectories, suggesting entropy is an effective critical-point selection metric.

\subsection{Empirical Validation of Uncertainty}
\label{sec:uncertainty}

An analysis of the detected points of uncertainty revealed a clear linguistic correlation. Tokens exhibiting the highest average entropy—such as \textit{``For''}, \textit{``Given''}, and \textit{``Since''}—frequently marked logical transitions, the introduction of sub-problems, or causal relations. Other high-entropy tokens, including \textit{``First''} and \textit{``According''}, were also observed to initiate new reasoning steps. These tokens correspond precisely to structural moments in reasoning where the model must choose among multiple plausible logical continuations.

\begin{figure}[b]
    \centering
    \includegraphics[width=0.6\columnwidth]{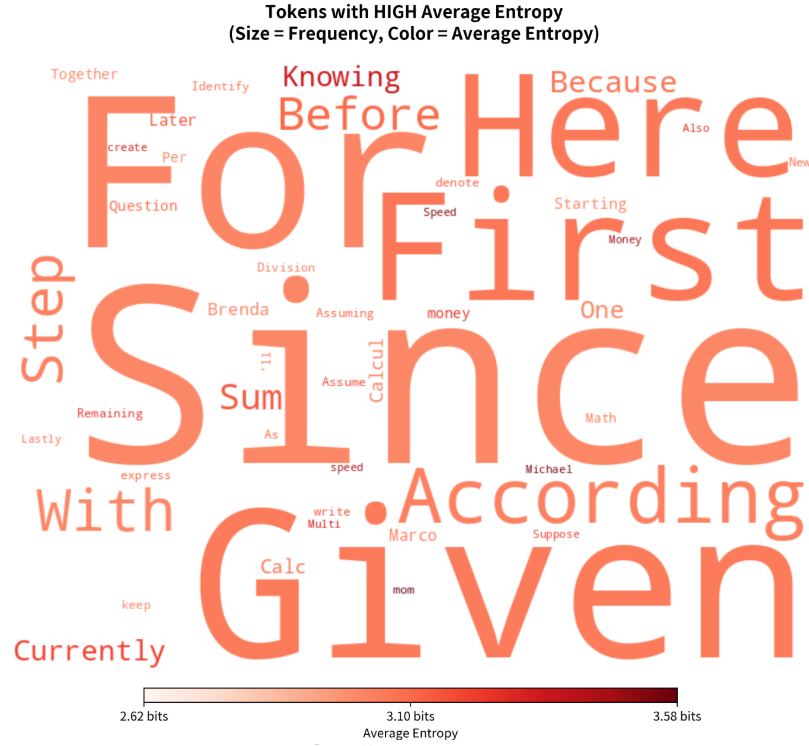}
    \caption{
    Word cloud of tokens with high average entropy observed in the \textit{SVAMP} dataset. 
    Token size represents frequency and color encodes the average entropy (in bits), highlighting terms such as 
    \textit{“For”}, \textit{“Since”}, and \textit{“Given”}, which are associated with logical transitions in reasoning.
    }
    \label{fig:entropy-wordcloud}
\end{figure}

At these critical junctures, the probability distribution over output tokens becomes notably dispersed, reflecting the model’s indecision regarding the most coherent semantic direction. 

These entropy spikes frequently coincide with discourse connectives (e.g., “for”, “since”, “given”), suggesting that the model’s uncertainty aligns with structural transitions in linguistic reasoning. Such correspondence supports the view that entropy can reveal moments of linguistic introspection, where the model implicitly evaluates competing semantic paths before proceeding.

This pattern supports the hypothesis that entropy functions as an indicator of \textit{local cognitive complexity}: points where fast, automatic generation (\textit{System 1}) gives way to slower, more analytical reasoning (\textit{System 2}). Thus, the entropy-based mechanism effectively monitors internal signals of cognitive hesitation, transforming them into opportunities for reflective refinement.

The distribution of high-entropy tokens is visualized through a word cloud (Figure~\ref{fig:entropy-wordcloud}), highlighting terms associated with logical transitions in reasoning. This linguistic correspondence suggests that the method captures internal traces of hesitation within the model’s own generative process, approximating a primitive form of introspection.

\section{Conclusion and Future Work}

PREGU successfully achieved its objective by developing an adaptive architecture that integrates partial reasoning generation in the token space with localized refinement in the latent space, guided by entropy. The approach validates the hypothesis that internal uncertainty can serve as a metacognitive control mechanism to direct computational effort toward regions offering the highest potential for informational gain.

\subsection*{Known Limitations}

PREGU inherits certain limitations from the original \textit{Soft Reasoning} approach. First, the scope of Bayesian Optimization remains restricted, as it operates only on the initial embedding following a detected uncertainty point, thereby constraining refinement to a localized region of the latent space. Second, the method’s reliance on the internal verifier ($r_{\text{verifier}}$)—which employs the LLM itself to assess the correctness of its own outputs—can introduce bias, particularly in complex mathematical verification tasks. Finally, the fixed entropy threshold ($\tau$) constitutes a sensitive hyperparameter: inadequate calibration may either cause premature fragmentation of reasoning sequences or reduce the frequency of uncertainty detection, effectively reverting the behavior toward that of the original Soft Reasoning method.

\subsection*{Future Work}

Future research should aim to expand the scope of optimization within the latent space to encompass a window of multiple embeddings following the point of uncertainty, rather than a single one. This broader search may enable smoother and more contextual refinements. Another promising direction involves designing new reward metrics that reduce dependence on the LLM's self-evaluation, for instance by integrating external or symbolic verifiers for mathematical reasoning. Additionally, dynamically calibrating the entropy threshold ($\tau$)—adapting it to the problem’s complexity and context—and quantifying explicit cost-benefit metrics (e.g., execution time, inference cost per correct solution) represent important steps toward improving PREGU’s efficiency, interpretability, and robustness.

%%
%% The acknowledgments section is defined using the "acknowledgments" environment
%% (and NOT an unnumbered section). This ensures the proper
%% identification of the section in the article metadata, and the
%% consistent spelling of the heading.
\begin{acknowledgments}
This work has been fully/partially funded by the project Research and Development of Digital Agents Capable of Planning, Acting, Cooperating and Learning supported by Advanced Knowledge Center in Immersive Technologies (AKCIT), with financial resources from the PPI IoT/Manufatura 4.0 / PPI HardwareBR of the MCTI grant number 057/2023, signed with EMBRAPII
\end{acknowledgments}

%% The declaration on generative AI comes in effect
%% in Janary 2025. See also
%% https://ceur-ws.org/GenAI/Policy.html
\section*{Declaration on Generative AI}
  
During the preparation of this work, the author(s) used ChatGPT (GPT-5.2) to assist with translation and language editing (grammar, spelling, and clarity). The author(s) reviewed and edited the output and take full responsibility for the final content.

%%
%% Define the bibliography file to be used
\bibliography{sample-ceur}

%%
%% If your work has an appendix, this is the place to put it.
\appendix
\section{PREGU - Experiments}

\subsection*{Experimental Setup and Reproducibility}

All experiments were executed on an NVIDIA DGX-H100 system equipped with 8× H100 GPUs (80\,GB each).

The models evaluated were \textbf{LLaMA-3-8B}, \textbf{Mistral-7B}, and \textbf{Qwen2-7B}, selected for their architectural diversity and open accessibility.
All experiments followed the same inference configuration, including token truncation to the top-k ($k=50$) logits for entropy estimation and a fixed entropy threshold of $\tau = 3.0$ bits.
Uncertainty detection followed $t_{\min}=5$ (minimum prefix), and we trigger at most one interruption per path (first token where $H_t \ge \tau$).

For the latent-space optimization phase, Bayesian Optimization was applied over a projected subspace of $d=50$ dimensions, using the Expected Improvement (EI) acquisition function. 
Each refinement round sampled $k=5$ latent perturbations per reasoning path, guided by the composite reward function in Eq.~\ref{eq:reward}.

To account for stochasticity in both token sampling and latent perturbation, each experimental configuration was executed across multiple independent runs with distinct random seeds. 
All results were reported as mean and standard deviation to capture the natural variability of large language model inference.
Benchmarks included GSM8K, GSM-Hard, SVAMP, and StrategyQA under the zero-shot setting, following the standard prompt structures described in the methodology.

For each benchmark, a random subset of 200 problems was uniformly sampled from the full dataset at the beginning of each run. 
This random selection ensured that different executions explored distinct subsets while maintaining comparable task difficulty distributions. 
The random seed controlling the sampling process was synchronized with the inference seed, guaranteeing consistency between data selection and generation randomness across runs.

Finally, representative examples from different benchmarks are included in the Appendix to illustrate the full execution trace of PREGU. 
Each example shows both stages of operation—entropy-based interruption and latent-space refinement—highlighting how uncertainty localization leads to structured reasoning improvements. 
These examples provide a transparent depiction of the reasoning process and facilitate independent verification of the experimental pipeline.

\medskip
\noindent\textbf{Note.}
The examples reported here correspond to raw model outputs. 
Therefore, they may include hallucinations or other generation artifacts (e.g., spurious claims or incoherent fragments), and are provided solely to illustrate the behavior of the proposed pipeline.

\subsection{PREGU Execution Example — StrategyQA}
\label{app:pregu-exec-372}

\noindent\textbf{Initial Question.}
\emph{Mayor: head of municipal government such as a town or city. Are Mayors safe from harm from the federal government?}

\medskip
\noindent\textbf{Ground Truth:} False

\subsection*{Phase 1: Partial Reasoning Generation}
\noindent (Entropy pause marked as \texttt{[STOP]}. Rewards refer to the partial prefix quality.)

\medskip
\noindent\fbox{\parbox{\linewidth}{
\textbf{Path 0}\\
\texttt{\small 1) Mayors are[STOP]}\\
}}
\medskip
\noindent\fbox{\parbox{\linewidth}{
\textbf{Path 1}\\
\texttt{\small 1) Mayor is a position of municipal government. 2) The municipal government,[STOP]}\\
}}
\medskip
\noindent\fbox{\parbox{\linewidth}{
\textbf{Path 2}\quad \emph{(Selected)}\\
\texttt{\small 1) The mayor is the head of a municipal government. 2) The federal government is a higher level of government. 3) The constitution grants[STOP]}\\
}}
\medskip
\noindent\fbox{\parbox{\linewidth}{
\textbf{Path 3}\\
\texttt{\small The Mayor is the local leader of a city or town. A federal government[STOP]}\\
}}
\medskip
\noindent\fbox{\parbox{\linewidth}{
\textbf{Path 4}\\
\texttt{\small 1) The Mayor is the leader of the local government and not part of the federal government. 2) No[STOP]}\\
}}

\subsection*{Phase 2: Complete Reasoning Chains}
\noindent (Each path is refined in latent space and yields a complete answer.)

\medskip
\noindent\fbox{\parbox{\linewidth}{
\textbf{Path 0 — Complete Chain}\\
\texttt{\scriptsize ... auge officials of municipal governments, which are part of the state government, not the federal government. 2) The federal government has no direct authority over municipal governments, but it can influence them through various means such as funding or laws. 3) However, the federal government does not have the power to harm mayors directly, as they are not federal officials.\newline Answer:\newline True. Mayors are safe from harm from the federal government.}\\[2pt]

}}
\medskip
\noindent\fbox{\parbox{\linewidth}{
\textbf{Path 1 — Complete Chain}\\
\texttt{\scriptsize ... oroften, has its own laws and regulations, but it is part of the larger national government. 3) The federal government has the power to enforce laws and regulations, and it can intervene in the affairs of municipal governments if necessary. 4) However, the federal government is not typically involved in the day-to-day operations of municipal governments, and mayors are usually free to carry out their duties without interference from the federal government. 5) Therefore, mayors are generally safe from harm from the federal government, but there are exceptions if the mayor is breaking federal laws or if there is a national emergency.\newline Answer:\newline Mayors are generally safe from harm from the federal government, but there are exceptions.}\\[2pt]
\textbf{Answer:} mayors are enerally safe from harm from the federal overnment but there are exceptions
}}
\medskip
\noindent\fbox{\parbox{\linewidth}{
\textbf{Path 2 — Complete Chain}\quad \emph{(Selected)}\\
\texttt{\scriptsize ... kh the power to the federal government to regulate the states and municipalities. 4) The federal government can use its power to investigate and prosecute individuals, including mayors, if they are suspected of violating federal laws. 5) However, the federal government is not allowed to arbitrarily harm or threaten mayors.\newline Answer:\newline False. Mayors are not safe from harm from the federal government if they are suspected of violating federal laws, but the federal government is not allowed to arbitrarily harm or threaten mayors.}\\[2pt]

}}
\medskip
\noindent\fbox{\parbox{\linewidth}{
\textbf{Path 3 — Complete Chain}\\
\texttt{\scriptsize ... iederates over the entire country, not just a single city or town. Therefore, the federal government has the power to intervene in the affairs of a city or town, including the removal or replacement of the Mayor. So, Mayors are not safe from harm from the federal government.\newline Answer: False.}\\[2pt]

}}
\medskip
\noindent\fbox{\parbox{\linewidth}{
\textbf{Path 4 — Complete Chain}\\
\texttt{\scriptsize ... dit: The Mayor is not a federal official, so they are not under the protection of the federal government. 3) The federal government has no direct authority over the Mayor unless there is a violation of federal law. 4) The Mayor is not immune from prosecution if they break federal law.\newline Answer:\newline False. The Mayor is not safe from harm from the federal government if they break federal law.}\\[2pt]

}}

\subsection{PREGU Execution Example — GSM8k}
\label{app:pregu-exec-463}

\noindent\textbf{Initial Question.}
\emph{Ashley's pizza delivery costs \$15. What is the total amount that Ashley should give the delivery man if she wants to give a tip that is equal to 1/5 of the amount she ordered?}

\medskip
\noindent\textbf{Ground Truth:} 18

\subsection*{Phase 1: Partial Reasoning Generation}
\noindent (Entropy pause marked as \texttt{[STOP]}.)

\medskip
\noindent\fbox{\parbox{\linewidth}{
\textbf{Path 0}\\
\texttt{\small 1/5 means Ashley should give 1/5 of the total price.[STOP]}\\
}}
\medskip
\noindent\fbox{\parbox{\linewidth}{
\textbf{Path 1}\\
\texttt{\small 1/5 of the ordered amount is the tip Ashley wants to give. So, to find the amount of the tip, we[STOP]}\\
}}
\medskip
\noindent\fbox{\parbox{\linewidth}{
\textbf{Path 2}\quad \emph{(Selected)}\\
\texttt{\small Ashley's tip should be 1/5 of what she paid for a delivery, so the tip amount is 1/5 * \$15.[STOP]}\\
}}
\medskip
\noindent\fbox{\parbox{\linewidth}{
\textbf{Path 3}\\
\texttt{\small 1/5 of Ashley's pizza delivery is her tip. So, to find the tip amount, we[STOP]}\\
}}
\medskip
\noindent\fbox{\parbox{\linewidth}{
\textbf{Path 4}\\
\texttt{\small 1/5 (twentieth) of the[STOP]}\\
}}

\subsection*{Phase 2: Complete Reasoning Chains}
\noindent (Each path is refined in latent space and yields a complete answer.)

\medskip
\noindent\fbox{\parbox{\linewidth}{
\textbf{Path 0 — Complete Chain}\\
\texttt{\scriptsize ... aille's pizza delivery costs \$15, so we need to find 1/5 of \$15.\newline Answer: \$3}\\[2pt]
}}
\medskip
\noindent\fbox{\parbox{\linewidth}{
\textbf{Path 1 — Complete Chain}\\
\texttt{\scriptsize ... RPC:\newline 1. Multiply the cost of the pizza by 1/5.\newline 2. Add the tip to the cost of the pizza.\newline \newline Answer:\newline \$16.50 (15 + (15 * 1/5))}\\[2pt]
}}
\medskip
\noindent\fbox{\parbox{\linewidth}{
\textbf{Path 2 — Complete Chain}\quad \emph{(Selected)}\\
\texttt{\scriptsize ... Now, to find the total amount Ashley should give the delivery man, we add the tip amount to the cost of the delivery: \$15 + (1/5 * \$15).\newline Answer: \newline \$18}\\[2pt]
}}
\medskip
\noindent\fbox{\parbox{\linewidth}{
\textbf{Path 3 — Complete Chain}\\
\texttt{\scriptsize ... Luxury Real Estate in Dubai will divide the cost of the pizza delivery by 5. Then, we will add the tip amount to the cost of the pizza delivery to find the total amount Ashley should give the delivery man.\newline \newline Answer:\newline \$15 + (\$15 / 5) = \$15 + \$3 = \$18.}\\[2pt]
}}
\medskip
\noindent\fbox{\parbox{\linewidth}{
\textbf{Path 4 — Complete Chain}\\
\texttt{\scriptsize ... imanager's cost is the tip. So, we need to find 1/5 of \$15.\newline Answer: \$3}\\[2pt]
}}

\medskip
\noindent\textbf{Selection Rule.}
Each partial reasoning sequence generated in \emph{Stage 1} serves only as an exploratory prompt. 
The final selection among reasoning paths occurs after the complete refinements in the latent space (\emph{Stage 2}), 
based on the final evaluation scores. 
The chosen path thus corresponds to the reasoning that achieved the highest overall reward according to the combined objective function.

\medskip
\noindent\textbf{Discussion.} 
The examples provided demonstrate the operational dynamics of PREGU, showing how uncertainty detection and localized latent-space optimization interact to enhance reasoning reliability. 
Together, they offer a transparent view of the model’s decision process and its transition from intuitive to analytical reasoning.

\end{document}